\pdfoutput=1

\documentclass[11pt]{article}

\usepackage[preprint]{acl}

\usepackage{times}
\usepackage{latexsym}

\usepackage[T1]{fontenc}

\usepackage[utf8]{inputenc}

\usepackage{microtype}

\usepackage{inconsolata}

\usepackage{graphicx}
\usepackage{enumitem}

\newcommand{\pquote}[2] {{\textit{``#1'' (#2)}}}

\title{Beyond Text: Characterizing Domain Expert Needs in Document Research}

\author{Sireesh Gururaja \quad Nupoor Gandhi \quad Jeremiah Milbauer \quad Emma Strubell\\
Language Technologies Institute, Carnegie Mellon University, Pittsburgh, PA, USA\\
\texttt{\{sgururaj, nmgandhi, jmilbaue, estrubel\}@cs.cmu.edu}
}

\begin{document}
\maketitle
\begin{abstract}








Working with documents is a key part of almost any knowledge work, from contextualizing research in a literature review to reviewing legal precedent. Recently, as their capabilities have expanded, primarily text-based NLP systems have often been billed as able to assist or even automate this kind of work. But to what extent are these systems able to model these tasks as experts conceptualize and perform them now? In this study, we interview sixteen domain experts across two domains to understand their processes of document research, and compare it to the current state of NLP systems. We find that our participants processes are idiosyncratic, iterative, and rely extensively on the social context of a document in addition its content; existing approaches in NLP and adjacent fields that explicitly center the document as an object, rather than as merely a container for text, tend to better reflect our participants' priorities, though they are often less accessible outside their research communities. We call on the NLP community to more carefully consider the role of the document in building useful tools that are accessible, personalizable, iterative, and socially aware.
\end{abstract}

\section{Introduction} 

From contextualizing scientific research in literature reviews, to understanding the functioning of complex organizations, experts conduct a wide variety of tasks that depend on document research. Document research, i.e. reading, understanding, and otherwise working with collections of documents is a process that underlies almost all knowledge work. As such, there is a rich body of literature that aims to understand how experts in various fields read documents, characterizing goals, processes, and how their experiences and knowledge inform what and how they read \citep[][\textit{inter alia}]{bazerman_physicists_1985, hillesund_digital_2010, mysore_how_2023}. 

More recently, primarily text-based NLP tools such as LLMs have been proposed as "solutions" to document research. General purpose commercial models are billed as being able to "understand" both single documents and even whole corpora, context length limits willing, and there are a growing number of purpose built tools targeted at particular professions, from legal document tasks 
\cite{harvey, luminance, donotpay}
to aiding in the process of scientific discovery \citep{ai4science_impact_2023}, with some claims going as far as to argue that some parts of the scientific process could soon be wholly automated  \citep{lu2024aiscientistfullyautomated}.

But to what degree are LLMs able to model document research and understanding as currently carried out by experts? In this study, we interview 16 domain experts working in materials science, law, and policy, to understand their processes of document research: Their goals, the uses they have for these documents, and how they evaluate the documents' content for relevance and quality.

In our analysis, we derive tasks common to the experts we interview, and assess the degree to which modern tools from NLP and adjacent fields address those tasks. We find that our experts' processes are highly personal and varied, involve iteratively constructing mental model, and are consistently informed not only by the content of the documents, but by the material and social context of their production. Tools that design around this context, such as citation-aware tools for scientific support \citep{he_paperpoles_2019, Heimerl2016CiteRiversVA}, better reflect experts' processes, though they tend to be limited to domains that explicitly center publication structure.

By contrast, modern, general-purpose systems from NLP tend to reflect a common, information-centric view: Documents are merely containers for information, and that information within documents can therefore be segmented, decontextualized and displayed without regard for the source document. This is reflected in the affordances of the systems themselves, which often operate either on individual sentences, or on segments whose lengths are determined by an underlying model's context length or a chunking algorithm \citep{asai-etal-2023-retrieval}, which may or may not align with semantic boundaries \citep{qu2024semanticchunkingworthcomputational}.

The experts we interviewed view documents in ways that cohere much more strongly with theory found in the Science and Technology Studies (STS) literature: that documents are not merely conduits for information, but are traces of social processes whose details are crucial to the documents' interpretation and use. For our experts, these details can inform evaluating a document's provenance, assessing a document against background knowledge of a field, or considering the global context in which a document exists. In  other words, a document ``serves not simply to communicate, but also to coordinate social practices'' \citep{brown_social_1996}. We therefore call for the NLP community to develop \textbf{accessible} systems and methodologies that are more \textbf{personalizable}, \textbf{iterative} and \textbf{socially aware} (\S\ref{section:conclusion}) in order to more accurately reflect the views and priorities of their users, and the rich social context in which documents are produced and consumed.


\section{Related Work}

\paragraph{Understanding Reading.} We locate this work in the tradition of work that seeks to understand expert readers and their processes for working with the documents they read, which often take the form of interview studies. \citet{mysore_how_2023}, for example, conducts semistructured interviews and think-aloud sessions with data scientists for how they conduct literature review. We also see many similarities with the findings in \citet{bazerman_physicists_1985}, who finds that physicists rely on "purpose-laden schemas", which include models of both content and authorship and other metadata, similar to what we find with materials scientists.

\paragraph{Document Theory.} We draw on the STS literature for theories of document-centric views of knowledge work. \citet{lund_document_2009} provides a broad overview of document studies beginning in the early 20th century, focusing on the materiality and social production of documents through the digital age. He points out the shift in focus away from documents towards disembodied information in library sciences in the 1960s. \citet{frohmann_scientists_2004} corroborates this shift (albeit with slightly different dates) and places it contemporaneously with ``discourses of...artificial intelligence and informatics,'' while arguing for an understanding of the contingent, social role of the scientific publication. These two sources illustrate the a possible origin for the elision of the document in the framing of contemporary NLP. \citet{brown_social_1996} similarly argue that a document, rather than being a ``conduit'' for information, serve as a mode of social coordination and control in their production and distribution. 

\paragraph{Document-aware Reading Support.} Document awareness as a principle of system design is an active area of research, especially in human-computer interaction and information retrieval. The Semantic Reader project \citep{lo_2024_semanticreader} incorporates a number of these features into an reader that shows an enriched view of a PDF document, and works like \citet{he_paperpoles_2019} and \citet{Heimerl2016CiteRiversVA} allows exploration through citations and other metadata. Work in NLP that accounts for metadata like citations, such as \citet{viswanathan-etal-2021-citationie}, is less common. We note, however, that metadata is seldom considered in reading support work outside of scientific documents, and personalized reading support, i.e. work that considers the context of the reader, is also rare.

\paragraph{Challenges to NLP.} There is a growing ambivalence in NLP towards the practice of benchmark-based evaluation \citep{gururaja-etal-2023-build}. This paper joins a growing number which call for benchmarks to be more closely aligned to end-user needs. \citet{newman-griffis-etal-2021-translational} call for what they term ``translational NLP,'' which proposes an application focus as the driver of scientific progress. \citet{katz-etal-2023-neretrieve} propose a new benchmark that presumes the strength of LLMs at traditional NER tasks as the basis for proposing a much more difficult benchmark that better reflects information seeking needs.

\section{Methods}

We recruited 16 participants, beginning with a convenience sampling method \citep{galloway2005sampling}, in which the authors began by interviewing existing non-computer science collaborators across the projects they worked on, and then by snowball sampling \citep{parker2019snowball}, in which participants were asked to recommend other interview candidates. Our participants, six women and ten men, were drawn from collaborations in the materials science, law, and policy communities, were all based in the U.S. and had a wide age distribution, with five participants between ages 25 and 34, six between 35 and 44, three between 45 and 54, and two 55-64. Ten of our participants were associated with materials science, though many identified themselves as belonging to other disciplines, such as chemical or mechanical engineering. All of our participants in this group were either professors or postdoctoral researchers whose primary focus was the synthesis, characterization, or modeling of materials. As such, the document research that they described to us was primarily the process of literature review: keeping up to date with the subfields they already worked in, or learning about new subfields that became interesting to their work. The remaining six participants were academics and professionals whose jobs involve reading, researching, interpreting, or otherwise engaging with law, public policy, or governmental records. This population's document research operated on many more kinds of documents, but the informational goals were largely similar, e.g. staying abreast of relevant policy, legal precedent, or public reactions to policy. 

These populations are neither very similar nor dissimilar; we use them as a way of understanding what themes in how experts work with documents might generalize across groups with nominally different tasks, and what themes might be specialized to a single domain. In essence, we aim to establish a loose lower bound of the variety of tasks that professionals across different domains carry out.

We conducted semi-structured interviews \citep{weiss1995learning} with our participants that lasted between 27 and 73 minutes, with the median interview lasting 53 minutes. Interviews were conducted with a dedicated notetaker, and we recorded the interview audio with participant consent. We followed an interview guide (included in Appendix~\ref{section:interview-guide}), that developed four broad themes: The participant's current work and positionality,  their current process for document research including how it fit into their work, how they evaluate documents for relevance and/or quality, and finally what existing tools they use to perform document research. We developed this set of questions in collaboration with materials scientists on one author's project, and was evaluated with a test interview before wider interviewing, with only minor changes to the wording of some questions. The guide did not change between the two subpopulations we interviewed. We conducted the interviews between February and July 2024. 

Following the procedures of grounded theory \citep{strauss1990basics}, at the conclusion of each interview, authors produced an analytical memo  detailing the themes from that interview \citep{glaser2004remodeling}. After a sufficient number of interviews for recurring, coherent themes to emerge, authors began a process of independently open coding the data, developing a thematic taxonomy that generalized across interviews. After this, the authors met regularly to discuss and refine the open codes into a preliminary closed coding frame~\citep{miles1994qualitative}. Authors then annotated each interview with the themes from this closed coding frame. In analysis meetings, the authors further refined this closed coding frame by adding, merging, and removing codes, then iteratively re-coding the data. The analysis of this paper emerged from closed-coded versions of the data, and was validated against the original transcripts for appropriate context.

\section{Summary of Interviews}
\label{section:overview}
Document research implied a wide variety of activities to our participants. Despite the variety of tasks, however, several common threads emerged that bridged the disciplinary gap. We conceptualize the tasks carried out by our participants across domains to broadly fall into three categories: Local context tasks, global context tasks, and corpus construction. In the following sections, we first characterize the logistics of the documents that our participants work with, describe some of the examples of each of these three types of task, as well as some broader themes across participants.

\subsection{Document Characteristics} 

The documents our participants worked with were primarily, though not always, in PDF format (with exceptions including maps and raw data files), and originated across a wide temporal range. Several participants described having to work with scanned documents which may or may not have had OCR applied to them. In the case of materials scientists, many often consulted technical reports from the 1950s to 70s that were originally typewritten; some of our policy experts looked at digitized government documents that were hand-scanned. Even in cases where documents had been produced digitally, understanding rich visual content, like page layouts, tables, and charts was a key area of focus for nearly all of our participants. 

\subsection{Task types}
    \label{section:task_types}

    \paragraph{Local context tasks.} Local context tasks, which only involve the content of a single document, resemble common information extraction tasks. In the materials science context, this often manifested as extracting information about how experiments were conducted, the materials that resulted, and their associated properties, in keeping with the principles of data-driven design of materials \citep{himanen2019data, olivetti2020data}. For instance, participant 21 described how their students \pquote{work on collecting information from the articles, and...build the models that can predict the material properties}{21}.
    Information extraction tasks were also present in the policy domain, with one participant describing \pquote{trying to extract policy data from these plans, including...different entities, different policy parameters}{1}. Local context tasks are only carried out once a researcher has already developed a mental model for the content of the documents they are searching, and are conducted only in corpora and subfields that the researcher knows well. 
    
    \paragraph{Global context tasks.} By contrast, global context tasks, which rely on signals from other documents, or the focus document's connections to them, are often much more exploratory. Common to all of our participants was the task of coming to understand a new corpus or subdomain with which they were previously unfamiliar; our experts were frequently reading and working with documents outside their core area of expertise. In materials science, this was described as \pquote{ building my own intuition of a classic material science thing...if I vary this, it goes up or it goes down.}{2}; in the law and policy domain, this could be understanding a corpus of government communications obtained through a freedom of information request, or the ramifications of a new policy through the public response to it. In these cases, information extraction-like approaches are insufficient: as one participant put it, \pquote{it wouldn't be...sufficient to say...we're just searching for a needle in a haystack...we're interested in understanding the haystack.}{39}. Researchers described constructing a mental model that was subdomain- or corpus-specific, progressively refining that mental model through the encounter with more documents. Over time, this allowed them to develop intuitions and expectations of the corpus, which also provided signal when they were subverted. This process parallels the background knowledge integral to evaluating the novelty and epistemic status for materials scientists' keeping up-to-date with their existing interests. Global context tasks were seen as a precursor to local context tasks: only after building a reliable mental model for an area did our participants feel comfortable looking at individual documents one at a time.
    
    \paragraph{Corpus construction.} While materials scientists consistently described a standard workflow that relied on academic search engines, these types of resources were only available to our participants in law and policy in the case of firm-internal documents or legal precedent. More commonly, participants explained that it was not trivial to collect corpora, beginning with \pquote{putting in phone calls to various libraries...to sort of find out what material is available}{39}. Our participants frequently described having assemble their corpora from documents \pquote{chopped up into different chapters}{8}, or that contained a \pquote{reference to some other document that contains the relevant information}{1}. Constructing corpora in this way often involves a great deal of expertise, both in knowing what to include, by conducting \pquote{manual verification of the documents we retrieve to understand if they actually are the documents we're seeking.}{1}, and verifying \pquote{the degree to which it is complete or extensive, that's an important consideration.}{1}. 

\subsection{Broader themes}

\paragraph{Awareness of contemporary technology.} We asked about experts' current processes, including the tools they had used, or whether they had incorporated AI tools into their workflow. While many of our participants were technically sophisticated, with some training their own BERT-based models to do topic classification, or writing an emacs-based tool for paper discovery, usage of LLM tools like ChatGPT was largely constrained to non-document tasks like writing or code assistance. By contrast, when attempting to use them for document research tasks, they described a number of pitfalls. Some expressed doubts about the lack of specific background knowledge, or concern about ceding control of the research process; others described trying to to get models to work for their process and facing challenges. One participant, for instance, said \pquote{can we just dump a bunch of PDFs into a GPT and get a summary? And it turns out it wasn't that easy. It wasn't like plug and play. But it also was showing some potential.}{17}. Though some of our participants had evaluated contemporary tools, they still considered them models that had to be customized, as in earlier machine learning, rather than useful as drop-in tools.

\paragraph{Access to technology.} Of the tasks that we heard described, many are the subjects of active research in the NLP and NLP-adjacent communities. However, very few of our participants had access to these technologies, primarily because only a few of them could or chose to in furtherance of their tasks. This was perhaps nowhere more evident than in the case of digitization and OCR, where one participant described scoping a project based on \pquote{whether the material is digitized or whether we're going to need to digitize it.}{39}, later discussing how a digitization system that considered each page a separate document prevented them from using keyword searches effectively: \pquote{ if we had reliable high quality text, and we had our document organization [taken] into consideration...then we could have used a keyword based search as like a candidate classifier.}{39}

\paragraph{Personalization.} Regardless of field, a consistent theme that we observed in our interviews was how idiosyncratic each researcher's process of document research was. There seemed to be no agreement, even within fields, on what makes a document relevant to a given search or what cues a researcher might use to assess the epistemic status of individual claims. We view this as a major challenge for NLP systems.

\section{Task Analysis}

In this section, we identify key tasks that our participants shared across fields, and compare the current state-of-the-art NLP tools with the needs that our experts outlined.

\subsection{Information Extraction}
\label{section:IE}
Traditional IE, where we tag entities of concrete, well-defined types, would be useful for many of our participants. 
However, our participants' needs were for concepts that would be specific to their research objectives (i.e. not available in off-the-shelf models) and difficult to define succinctly, like tagging spans that provide evidence of 20-25 core political values in an argument, like \pquote{equality and justice, liberties, security, safety}{42}. Participants also described the acceptable granularity of extracted information varying per-project: \pquote{we just used like more bag-of-words-based, keywords-based approaches...but the whole research project in that case assumed a level of bluntness that wouldn't be appropriate for other projects.}{39}
This is different to standard benchmark datasets for IE where the
types are more concrete and well-defined \citep{ding-etal-2021-nerd, tjong-kim-sang-de-meulder-2003-introduction}. Further, supervised neural systems require a large amount of expensive annotations for each new tag set \citep{li2020survey}, though recent work on few-shot IE with LLMs has aimed to reduce the potential annotation burden \citep{hofer2018few, ashok2023promptner, huang2020few}.

Our participants also emphasized that not all IE tasks of interest to them involve only local information, conflicting with traditional IE focused on within-document and short document settings. They describe settings where IE would be applied to long scientific papers and policy briefs: \pquote{I don't think there's a way for me to get the information that I need without having the full document, but once I have that document, I will only use the specific portions that I need.}{11}. These documents may exceed the context window of LLM-based IE systems and result in poorer performance \citep{dagdelen2024structured}. 

Evaluating IE tools extrinsically can reveal the significance of existing vulnerabilities. The value of better aligning evaluation principles with user needs was highlighted by one of our participants: 
\pquote{
it would still dramatically reduce the amount of time a researcher would need to spend...
But the conclusion was that these tools...
were inadequate on their own}{39}
For example, \citet{adams2024longhealth} shows that LLMs do not perform well enough on long document clinical notes for reliable clinical use in question-answering. 
In relation extraction, there has been work to soften evaluation metrics to accommodate for the use of generative models, reflecting a shift towards evaluation methods that are aligned with downstream utility where it is often enough to recover spans with overlapping span boundaries \citep{jiang-etal-2024-genres}.
These types of evaluations would better highlight the most impactful open problems in IE.

\subsection{Multimodality and OCR}
\label{section:multimodality}
Understanding visual and layout features of documents was a priority for nearly all of our participants, the vast majority of whom worked with documents prominently featuring tables, charts, or which conveyed information through layout. One materials scientists characterized their needs as \pquote{"looking for a statement backed up by data and the data can be just numbers. It can be graphs. It can be pictures of microstructures or just all}{0}, and another described extracting \pquote{a ton of data and figures...because they're usually X, Y plots.}{5}.  These tasks are inherently multimodal: while some LLM approaches recommend linearizing tables into text, charts, images, and layout information must be handled visually. Compounding this is the tendency for these documents to be scanned or photocopied instances of paper documents, but which still necessitate the processing detailed above, resulting in it being \pquote{tough to really search through them, and so they might not...appear when you're doing a lit review on Google Scholar}{2}.

While visual understanding of documents is not a solved problem, dedicated multimodal layout understanding models like the LayoutLM series \citep[][\textit{inter alia}]{huang2022layoutlmv3}, which serve both as visual segmentation models as well as representations for downstream document reasoning tasks remain an active area of research, general-purpose models like Qwen-2-VL\citep{Qwen2VL} and LLaVA-Next \citep{liu2024llavanext} now include visual document understanding and OCR benchmarks like DocVQA \citep{mathew2021docvqa} and TextVQA \citep{singh2019textvqa}, and proprietary models like Claude now have modes designed explicitly around PDF processing \footnote{https://docs.anthropic.com/en/docs/build-with-claude/pdf-support}. 

\subsection{Iterative Search and Exploration}
\label{section:search}
One common theme repeated across multiple participants was that the search process is inherently iterative.
Rather than rely on a single set of results identified for a particular information need, researchers will iteratively expand their search across multiple stages, using results to inform each successive step, progressively building a mental model of the search space. 

Researchers iterate for different reasons.
Some seek to identify the provenance of the information they find: \pquote{I go and read a current paper and I find where they cited that they got information from. And I read where that person got the information from and I read where that person got the information from. I usually try and find the original sources to everything.}{5}, with one even stating \pquote{sometimes I read a paper and the main use of it is the references in the paper}{4}.
Others iterate precisely to establish the global context surrounding a particular paper: \pquote{you have to like build up this context around the paper. What came before it? What is it citing? And how does the content of that paper relate to the ideas that came before it?}{4}.

Iterative approaches do exist within the information retrieval literature. Initial retrieved documents can be used as a source of information on related lexical terms \cite{attar1977local}, which additional can help address issues like terminology drift (section \ref{section:terminology}), and the process of learning as you search, where \pquote{reading and coding...helps you generate, develop contextual knowledge}{39}. More recent work has focused on iterative approaches which search through the structure of documents \cite{min2019knowledge, zaheer2022learning, hsu2024retrieval}, albeit with a focus on the task of multi-hop reasoning \cite{yang2018hotpotqa} rather than information exploration.

Iterative construction of mental models could also be supported by the task of ontology induction.
While there has been recent work to induce concepts from individual documents \citep{matos-etal-2024-towards}, our participants highlighted the need for concepts detected at the corpus level that are tailored to a specific research question. Despite initial work in the scientific domain \citep{katz2024knowledge}, abstracting concepts across documents has been shown to be challenging for state-of-the-art LLMs \citep{guo-etal-2024-nutframe}, even without the per-question adaptation.

\subsection{Terminology}
\label{section:terminology}
One of the consistent difficulties that our participants faced across fields, similar to those in \citet{mysore_how_2023}, was terminology. Our participants described three types of terminology shift. The first, temporal, is when the meaning of a word or a term drifts over time. This might be because of shifts in community usage, like one materials scientist pointed out, 
\pquote{in like 2008-ish, the Chinese community decided that the word shock [testing] also means high rate Kolsky bar testing}{5}. It could also be because the referents of the words themselves have changed, as in the case of committee organization in a local government: \pquote{if you're looking for committee reports from before 2018 and you're looking for hospitals, the committee on hospitals didn't exist until 2018.}{40}. The second, domain-specific change, is when different fields use different words or terms for very similar things: \pquote{I call it a surrogate model. If you ask a statistician what they're going to call it, they're going to call it an emulator. If you ask someone in the reliability community...they're going to call it a response surface.}{17}. These two types of terminology shift are partially addressed in the existing literature. \citet{periti_lexical_2024} survey approaches to track new usage and senses for existing vocabulary; \citet{lucy_words_2022} quantifies domain-specific terminology usage and synonymy across fields; \citet{head-2021-scholarphi} provide context-sensitive definitions of technical terms and mathematical symbols. 

However, while existing systems work on either identifying terms as near-synonyms  or providing definitions, our participants emphasized that understanding the differences in meaning and why one term might be used instead of another was also important. For example, one materials scientist outlined how while ``oxidation'' and ``aqueous corrosion'' meant similar processes, the keywords used to search for one vs the other, and the numbers that would characterize those properties would be different: \pquote{if it's aqueous corrosion...they might care about the atmosphere or basically the liquid concentration a lot more...But if you were then going to oxidation in high temperature, they are mostly looking at mass gain data.}{2}  The simultaneous focus on the similarities and differences in mostly-synonymous words reflects a process in which our participants tended to jointly model the semantic content of documents alongside the documents' social context, like its authorship or intended audience. This was especially true with the final type of terminology shift, political, where people describe similar things, but may use different language to convey different valence, or signal which aspects of an issue are being prioritized, as one participant gestured to in the case of climate policy: 
\pquote{rural communities will talk about micro grids, not as a part of climate action, but as a way to get off the dependency on investor owned utilities like PG\&E.}{8}. 

In these cases, our participants not only had to engage in a iterative process to find the synonyms, they also had to reason about the positionality of the authors and why they might use a different term. Understanding identity, its presence in corpora, and its interaction with LLMs is still a new area of research: \citet{kantharuban2024stereotype} and \citet{li2024chatgpt} demonstrate the sensitivity of models like ChatGPT to implicit markers of identity in responses and refusals, respectively, and \citet{lucy-etal-2024-aboutme} investigated author positionality and community belonging, and \citet{milbauer-etal-2021-aligning} demonstrated cross-community lexical differences linked with ideology. Reading support that addresses the needs of our participants would need to unify several running themes of work in an accessible interface: providing definitions, enriching those definitions with understanding of where assumptions and practice across communities might differ and the implications of different terminology across domains.

\subsection{Corpus Construction}
\label{section:corpus_construction}
For corpus construction, which we describe in section \ref{section:task_types}, our participants most often described starting with collections of library resources and Google searches, only eventually moving to write web scrapers if the kinds of documents were similar enough. Automated tools, such as custom scrapers that use LLMs to explore documents \citep{huang_autoscraper_2024, ma2023laser} would be extremely useful for expanding corpora beyond what is feasible to manually collect. However, our participants stressed the need to reconstruct documents from chapters, understand document versioning, and how to iteratively build \pquote{some checks of what's missed by [a search], false positives, false negatives}{42} in constructing corpora, implying that document scrapers for constructing corpora would need to accommodate an iterative, exploratory style to truly function for this purpose.



\subsection{Global and Social Context}
\label{section:documents_social}
When discussing how they handled the uncertainty of dealing with potentially contradictory information from multiple documents, our participants repeatedly made clear that content alone was rarely sufficient either to search for documents, or to assess the results of such a search, with one participant stating that document content accounted for \pquote{at best...40 to 45\%}{40} of what made a document trustworthy. In order to determine the degree to which statements and document content could be trusted, our participants relied on a number of complementary contextual signals. In this section, we discuss the signals our participants used when working with corpora. We note, however, that most of the issues outlined in this section remain underexplored in language technologies; as such, we primarily describe desiderata for future efforts.


Perhaps the most commonly used signal for our participants was consistency checks against contextual knowledge of their field and ``standard practices'' within it. Even in the domain of scientific publications, where published, peer-reviewed papers are commonly taken as ``vetted and approved by the scientific establishment''\citep{cronin_citation_1984}, materials science participants reiterated the need to critically evaluate papers and their results based on \pquote{how well rooted [they are] on the first principles of the field,}{28} and whether a given document passed their notion of a  \pquote{self-consistency test}{0}, asking \pquote{Does this graph support comments and assertions made in the text?}{0}, even though they were the intended audience of the paper. If numerical results seemed unlikely (by intuition), researchers saw this as an indicator that a paper required additional scrutiny: \pquote{it looks like they're kind of cherry picking their data to make it look good. And you want...full disclosure. Are they highlighting cases where maybe the thing doesn't work?}{17} These standards were also seen as changing over time: one materials scientist pointed out that \pquote{These days, titanium alloys are used for very specific applications, which is then going to be biasing the data that's collected on them}{28}, and a policy researcher noted that \pquote{planning history is full of these best practices, which turn out to be failures 10, 20 years later.}{8} This poses a challenge to current NLP systems, which do not model the "unexpectedness" of a particular statement or document within the context of similar documents, instead treating most content in documents as propositional and uniform in importance, making them vulnerable to spin \citep{yun_caught_2025}.
\citet{liu2023anchor} and \citet{milbauernewssense} have recently developed approaches for linking information across documents, however the field still works within a framework of resolving conflicts in propositional knowledge (see e.g. \citet{xu-etal-2024-knowledge-conflicts}, who survey approaches to mitigating knowledge conflicts in models and conclude that current approaches remain insufficient), rather than dealing with the contingent, temporally and socially bound truth values found in documents.

Participants also often used metadata like authorship as part of assessing document reliability across all fields. However, we found that the signals that participants chose to rely on were highly individual, with two participants rarely agreeing on how to evaluate a document. We heard contradictory opinions on the value of metadata like publication venue, citations, the principal investigator of the lab group that published the paper, and even the quality of the diagrams and charts within the paper from materials scientists. Even in cases where a participant used a certain signal, they emphasized its contextual nature. In the case of citations, for example, the age of the subfield was an important qualifier for one participant, who said \pquote{There's an enormous variation in citation rate depending on the field and how specialized it is. So anything that's old and highly specialized like steel research, numbers are tiny. There's just so few people active.}{0} Both of these signals — background knowledge and metadata — can be seen as authors using a mental model of how their field functions in order to evaluate a document. Further, the document is the unit of analysis that allows both of signals to be used - consistency checks operate at the level of individual document components interacting, and the document is the unit which can be published, cited, and so on. Document-aware models and tools tend to be restricted to science, where metadata pertaining to documents is readily available through resources like S2ORC \citep{lo-etal-2020-s2orc}, but the necessity for document-based understanding is not: signals from the social context of document production like authorship and provenance were also important for our law and policy participants. 

Having a social model of the production of a document is even more important for our participants in cases where they are not the original or indended audience for a document, such as when examining planning documents, or communications obtained through government transparency programs. In these cases, participants often report having to piece together the information they seek by comparing multiple documents, including differences between versions of the same document, to get a better picture. For example, one participant details inferring political positioning for potential appointments in local government by \pquote{looking for a pattern of how someone may not be saying like their views, but in actuality are sort of expressing it through something that is legally binding.}{40} Another participant detailed a case in which the differences between versions of a document indicated important information about the honesty of the information: 
\pquote{they write a first draft, and then they literally edit out some of the material that's too honest for public consumption...I can't treat both of those pieces of information as equally important...the first one I know is more forthcoming, and the second one is more whitewashed.}{39}, where \pquote{a concept that's articulated infrequently relative to other concepts can be much more important for understanding the, like, underlying story.}{39}. As discussed in \ref{section:terminology}, modeling the types of social processes underlying document production is still new to NLP. Further, while document similarity is a robust and well-studied topic, finding meaningful differences in documents that might indicate information being hidden is much less common, with some work being published in legal NLP venues \citep{li_detecting_2022}.

\section{Conclusion}
\label{section:conclusion}
In this paper, we interview 16 domain experts from the domains of materials science and law/policy to understand how they conduct document research. We find that their self-described processes rely on understanding and actively modeling the social processes by which text is produced, through the construct of a document and its associated metadata. As exemplified by the success and usage of layout understanding models and the HCI work on scientific reading support, document-aware tools match the processes of our experts. More concretely, however, there are four qualities of systems that we call on the NLP community to build into new document research tools for expert users outside the field:

\paragraph{Accessible} Though we have characterized the state of existing research as it addresses many of the concerns of our experts, we note that the most accessible NLP tools today by far are primarily text-based and rarely consider documents a first-class citizen. By contrast, NLP work that does center documents is far less accessible. Many of the issues faced by our participants, whether working with older, undigitized documents, understanding terminology drift, or leveraging metadata to assess the provenance of a document are already the focus of existing research, but are not as accessible to them as web-based LLM systems. How can we both promote better modeling of how users engage with large quantities of text in accessible systems like LLMs, and make tools that do this modeling more accessible? 

\paragraph{Personalizable} We note in several places that our experts had idiosyncratic processes for finding, evaluating, and reading documents. From looking for different types of information both within and across fields, to having vastly different heuristics for assessing the provenance and reliability of a document, our participants, all successful experts in their fields, conduct research in highly personal, specific ways. Amid concerns of LLMs potentially stifling creativity and serving as a homogenizing force ~\citep{anderson_homogenization_2024, kumar_human_2024}, NLP tools should encourage and support diversity of thought by allowing for experts to customize systems to their existing personal processes.

\paragraph{Iterative} Almost all of our participants discussed the process of constructing a mental model as a series of iterative updates as they read new documents and reconciled the content with their expectations. This process of constructing a mental model was essential, and heavily used global context: experts relied extensively on signals like how any given document related to background knowledge, other documents in the corpora, standard practices in the field, as well as social signals like authorship and author positionality. As they read, their understandings of both the signals and how they pertained to the documents co-evolved. By contrast, NLP tools are often presented as either static or occasionally updated artifacts. To better support how experts work, NLP systems should malleably support their users' evolving mental models, including through easy schema updates, flexible data relabeling, and user-friendly retraining and evaluation loops.

\paragraph{Socially aware} The social character of document production played an important role for our participants, whether in understanding terminology and why it might be used, modeling authors and participants in spheres of discourse they participated in, and evaluating information for reliability and provenance that included information beyond the propositional content of the text they worked with. Our participants frequently looked beyond what was written in text to how and why it was written, treating documents more as traces of social processes than containers or conduits for pure, disembodied information. NLP tools should be designed around the idea that social context is crucial for understanding text: authorship, audience, communicative intent, and format are all factors that readers already consider; as readers already know, models should be aware that no text is \textit{just} text. 

\section*{Limitations}

Our interview study was evaluated and approved by an Institutional Review Board as STUDY2023\_00000431.

While we constructed our sample deliberately to span two fields, we acknowledge that this was a convenience sample, and that it is neither specific to one field, nor representative of all the ways that people might work with documents. Our study also focuses exclusively on potential users of NLP tools who are already experts in their domains. Their concerns are not likely to be representative of non-domain expert users, especially non-experts reading technical language.

\section*{Acknowledgments}

Research was sponsored by the Army Research Laboratory and was accomplished under Cooperative Agreement Number W911NF-22-2-0121. The views and conclusions contained in this document are those of the authors and should not be interpreted as representing the official policies, either expressed or implied, of the Army Research Laboratory or the U.S. Government. The U.S. Government is authorized to reproduce and distribute reprints for Government purposes notwithstanding any copyright notation herein.

\bibliography{custom, zotero_references, anthology}

\appendix

\section{Interview Guide}
\label{section:interview-guide}

Begin by defining document research: we’re interested in processes you have for finding documents, or things within documents to help you in research tasks in a professional context. 

\textbf{Demographics/Positionality}
\begin{enumerate}
    \item Can you briefly describe your job, covering the kinds of research questions that you encounter in your line of work? 
    \item In your work, can you describe the cases where you have to look for documents, or things within documents? An example would be great.
\end{enumerate}

\textbf{How do you do document research now?}
\begin{enumerate}[resume]
    \item Can you describe the goals that you have when you do document research? What kinds of documents and information do you search for? If you have different kinds of searches with different goals, please describe them.
    \item When searching for documents, are you searching for documents as a whole, or specific pieces of information/facts within those documents? How much of the document’s content as a whole do you end up using?
    \item What purpose do those documents or facts have once you find them? (reference? Quotation material? Prior approaches to what you’re trying to solve?)
    \item Is there an existing ontology to the kind of searching that you do? Are the things that you search for in documents part of a well-defined set of things, or is your approach to these documents creative?
    \item To what degree is finding documents or facts an iterative process? Is there a mental model that you have of the space of possible documents that you update as you find new documents?
    \item To what extent is the structure of documents relevant?  
\end{enumerate}

\textbf{How do you evaluate the documents that you find?}
\begin{enumerate}[resume]
    \item When you search for documents or facts, what does it mean for a document or fact to be high quality to your purpose? 
    \item When evaluating a document or fact for relevance or quality, how much of that depends on the content of the document itself?
    \item How much of that depends on knowledge of the field that you have that’s not explicitly in the document - other important documents, standard practices, etc? Are there resources for that kind of domain knowledge?
    \item How much depends on metadata, like citations, author affiliations, venue, etc?
\end{enumerate}

\textbf{Existing tools}
\begin{enumerate}[resume]
    \item What tools do you currently use to search for documents?
    \item When executing a search, how quickly do you usually find the sort of thing you’re looking for? Are there specialized keywords that get you to what you’re looking for? 
    \item If you use specialized tools for your domain, what do they do differently from generic-domain tools, like Google search? 
    \item Do you ever write code to enable better searching? What are the tasks that code helps you with that existing tools are insufficient for?
    \item If there was something that your tool could do differently or better, what would it be?
    \item Have you used AI-based tools to aid in your work? How well have they suited your workflow and process?
\end{enumerate}

\textbf{Demographics}
\begin{enumerate}[resume]
    \item I am going to read some age brackets. Can you indicate when I read a bracket that your age falls into?
    \begin{itemize}
        \item 18-24
        \item 25-34
        \item 35-44
        \item 45-54
        \item 55-64
        \item 65+
    \end{itemize}

    \item Is there anything else in your background that you consider relevant?
\end{enumerate}

\end{document}